\documentclass[a4paper, 10 pt, conference]{ieeeconf}
\IEEEoverridecommandlockouts 
\usepackage{amsmath,amsfonts}
\usepackage{algpseudocode}
\usepackage[ruled, vlined, linesnumbered]{algorithm2e}
\usepackage{array}
\usepackage{textcomp}
\usepackage{stfloats}
\usepackage{url}
\usepackage{verbatim}
\usepackage{graphicx}
\usepackage{caption}
\usepackage{subcaption}
\usepackage{adjustbox}
\usepackage{cite}
\usepackage{tabularx}
\usepackage{booktabs}
\usepackage{bm}
\usepackage{hyperref}

\title{\LARGE \bf
From Transparent Labware Segmentation to Collision Avoidance:\\ A Real-Time Edge-Aware Perception Pipeline
}
\author{
Shijun Ding$^{*}$,
Chen Qian$^{*}$,
Weiwei Shang,
and Junlin Xiong$^{\dagger}$%
\thanks{$^{*}$Shijun Ding and Chen Qian contributed equally to this work.}%
\thanks{$^{\dagger}$Corresponding author: Junlin Xiong.}%
}

\begin{document}

\maketitle
\thispagestyle{empty}
\pagestyle{empty}

\begin{abstract}

This paper presents an edge-aware instance segmentation framework that enables real-time robotic collision avoidance with transparent laboratory glassware using purely visual perception. Transparent vessels defy conventional segmentation due to refraction, specular reflection, and the absence of stable interior texture, yet their boundary contours remain comparatively reliable visual cues. Exploiting this observation, we augment a one-stage real-time instance segmentation backbone with a lightweight edge-detection branch, edge-guided attention fusion, and a parameter-free SimAM module, and further construct \textit{LabGlass-IS}, a 3{,}485-image, 21-category instance segmentation dataset of real laboratory glassware. The enhanced model achieves the highest Boundary F-score of 97.80 among compared methods, outperforming the YOLO-prompted FastSAM framework by 18.93 BF points. Furthermore, it maintains an inference speed of 7.1\,ms per frame and requires only 2.85\% of the parameters of the closest accuracy competitor. Multi-view triangulation of mask centroids further provides 3D positions for conservative bounding-volume collision constraints. Real-robot trials achieve a 93.3\% collision avoidance success rate, indicating the feasibility of the proposed perception-to-action pipeline for robot collision avoidance among fragile transparent objects. Our code is available at \url{https://github.com/havishamy/TransYOLO_3D}. Our video is available at \url{https://havishamy.github.io/paper-videos/}.

\end{abstract}

\section{Introduction}
Transparent laboratory vessels—such as beakers, test tubes, and Erlenmeyer flasks—are indispensable tools in chemical laboratories, serving as the primary carriers for sample storage, reaction execution, and quantitative analyses. These objects often coexist with robots, requiring reliable perception for safe motion planning and collision avoidance\cite{review}. Robotic perception systems commonly rely on depth sensors such as RGB-D cameras or LiDAR\cite{lif,lidar}. However, due to optical properties such as transparency, refraction, and specular reflection, these sensors frequently suffer from missing depth or measurement distortions when observing transparent objects\cite{cm,lif}. Stable perception of glass objects remains a challenging problem for robotic vision systems.

\begin{figure}[!t]
    \vspace{2mm}      
    \centering
    \includegraphics[width=86mm]{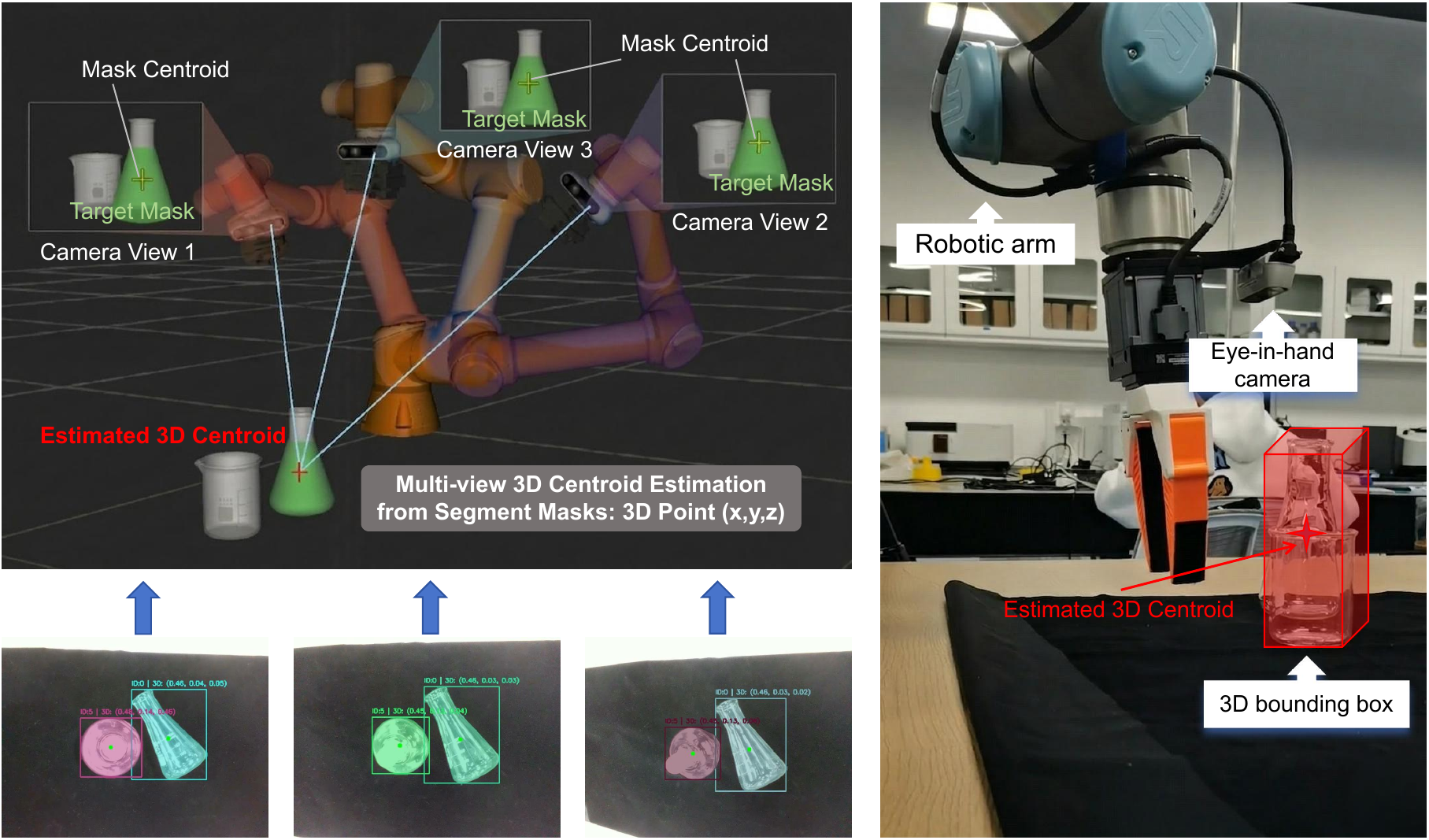}

    \caption{\textbf{Pipeline of perception-driven robotic collision avoidance for transparent labware.} The eye-in-hand camera on the robotic arm captures visual data, the 3D centroid of each vessel is derived from the segmentation mask centroid via multi-view triangulation, and a conservative 3D bounding box centered at the centroid is built as the collision avoidance constraint for robotic manipulation.}
    \label{fig_1}
\end{figure}

Recent studies have proposed specialized architectures\cite{lbsnet,trans2seg} for transparent object perception. CNN-based methods such as LBSNet\cite{lbsnet} enhance boundary-aware features through multi-scale fusion and additional refinement branches, which improve segmentation accuracy but introduce extra feature extraction and fusion operations that increase inference latency. Transformer-based approaches\cite{trans2seg} rely on global self-attention and multi-scale decoding, leading to substantial computational overhead and memory consumption due to dense attention operations. Foundation models like SAM\cite{sam} further improve generalization but require heavy encoders and large-scale feature processing, making real-time deployment on robotic platforms difficult. More importantly, most methods focus on semantic or instance segmentation accuracy on benchmark datasets, without converting perception results into spatial representations that can be directly used for motion planning or collision avoidance.

Another critical bottleneck is the scarcity of suitable datasets. Existing transparent object benchmarks \cite{translab,cleargrasp,cabd} primarily emphasize semantic segmentation or object detection, yet these annotations are insufficient for robotic manipulation, where precise separation of individual object instances is essential. Laboratory-oriented datasets such as LabPics \cite{labpics} offer only coarse category labels and exhibit substantial intra-class variability, which hinders the learning of fine-grained vessel geometry, structural details, and boundary cues. Furthermore, many large-scale datasets are synthetically generated \cite{cleargrasp}, and thus fail to faithfully reproduce the intricate optical phenomena, cluttered layouts, and illumination variations encountered in real chemical laboratories. As a result, models trained on such data often suffer from limited robustness and poor transferability in practical robotic settings, underscoring the urgent need for a dedicated real-world instance segmentation benchmark for transparent laboratory objects.

The core challenges can be summarized into three aspects. 
First, transparent glass objects exhibit complex optical phenomena such as refraction, reflection, and weak texture, which lead to unstable visual cues and make reliable detection and segmentation highly challenging. 
Second, many existing segmentation approaches are computationally heavy or designed purely for offline evaluation, making them difficult to deploy in real-time robotic manipulation and collision-avoidance systems, where an effective perception-to-action pipeline is required. Third, there is a clear shortage of real-world instance segmentation datasets for transparent laboratory glassware, while existing datasets often provide coarse annotations, limited instance-level labels, or rely heavily on synthetic imagery.

In response to the above challenges, this work contributes in perception, system integration, and dataset construction:
\begin{itemize}
\item At the perception level, we propose a lightweight edge-aware instance segmentation framework for transparent glass vessels, which explicitly strengthens boundary features to achieve accurate real-time segmentation.
\item At the system level, we construct a complete robotic perception and collision-avoidance pipeline (Fig.~\ref{fig_1}) by combining the proposed segmentation framework with multi-view 3D centroid estimation and conservative bounding-volume modeling.
\item At the data level, we introduce \textit{LabGlass-IS}, a real-world benchmark for transparent laboratory glassware instance segmentation, comprising 3,485 images, 21 categories, and 6,099 instance annotations.
\end{itemize}

\section{Related Work}

\subsection{Transparent Object Segmentation}

Transparent object segmentation has attracted increasing attention in recent years. Conventional appearance-based segmentation methods~\cite{yolact,yolo, yolactplus,solo,Mask2Former} often struggle to capture stable visual cues for transparent objects, leading to degraded performance in complex environments.

CNN-based approaches have been proposed to enhance boundary and contextual features. Translab~\cite{translab} jointly predicts masks and boundaries, while LBSNet~\cite{lbsnet} introduces dynamic boundary-aware feature fusion to improve segmentation quality. Transformer-based methods such as Trans2Seg~\cite{trans2seg} further strengthen global context modeling through self-attention and multi-scale decoding, but usually introduce higher computational overhead.

Foundation models such as SAM~\cite{sam} and its lightweight variant FastSAM~\cite{fast} have shown strong cross-task generalization. However, their performance on transparent laboratory glass vessels remains limited, and their computational overhead still challenges real-time robotic deployment. Overall, existing methods primarily emphasize segmentation accuracy, while efficient and boundary-aware instance segmentation for robotic applications remains unresolved.

\subsection{Vision-Based Robotic Manipulator Collision Avoidance}

Vision-based collision avoidance relies on reliable obstacle modeling for motion planning and safe manipulation. Early approaches typically employ precise 3D reconstruction or dense point cloud modeling, which are computationally expensive and sensitive to perception noise.

To enable real-time operation, recent studies adopt conservative geometric modeling strategies~\cite{collision,predictive}, representing obstacles with simple primitives such as bounding boxes or convex polyhedra for efficient collision detection. Deep learning-based methods further estimate object locations from RGB or RGB-D images to support fast and reliable planning~\cite{vision_based}.

For transparent objects, unreliable depth and optical effects make spatial estimation more challenging. Recent works improve geometry estimation through multi-view constraints~\cite{mvtrans}, refractive cues~\cite{rftrans}, and sim-to-real depth completion~\cite{cagt}. TORM~\cite{torm} reconstructs multiple transparent objects from multi-view masks for robotic grasping, while HEAPGrasp~\cite{heapgrasp} combines segmentation-based shape estimation with active hand-eye perception. Yet achieving stable and real-time obstacle representation using pure vision remains a challenging open problem.

\subsection{Chemical Laboratory Instrument Datasets}

Datasets play a critical role in transparent object perception. Early datasets such as LabPics~\cite{labpics} and CABD~\cite{cabd} are limited in scale, category diversity, or annotation precision, restricting their applicability in complex laboratory environments.

Recent datasets including Trans10K~\cite{translab} and ClearGrasp~\cite{cleargrasp} provide larger-scale data, but most focus on semantic segmentation or synthetic scenes and lack fine-grained instance annotations for laboratory glassware. Consequently, existing datasets are still insufficient for instance-level perception and robotic manipulation tasks in real chemical environments.

To address this gap, this work constructs a real-world instance segmentation dataset (\textit{LabGlass-IS}) for transparent glass laboratory vessels, containing 6,099 annotated instances across 21 common container categories.

\section{Method}

The proposed method consists of two main components: a lightweight edge-aware instance segmentation framework and a multi-view 3D centroid estimation pipeline with conservative bounding-volume modeling.

\subsection{Edge Detection and Edge-Guided Feature Module}

\begin{figure*}[!t]
\centering
\includegraphics[width=170mm,height=85mm]{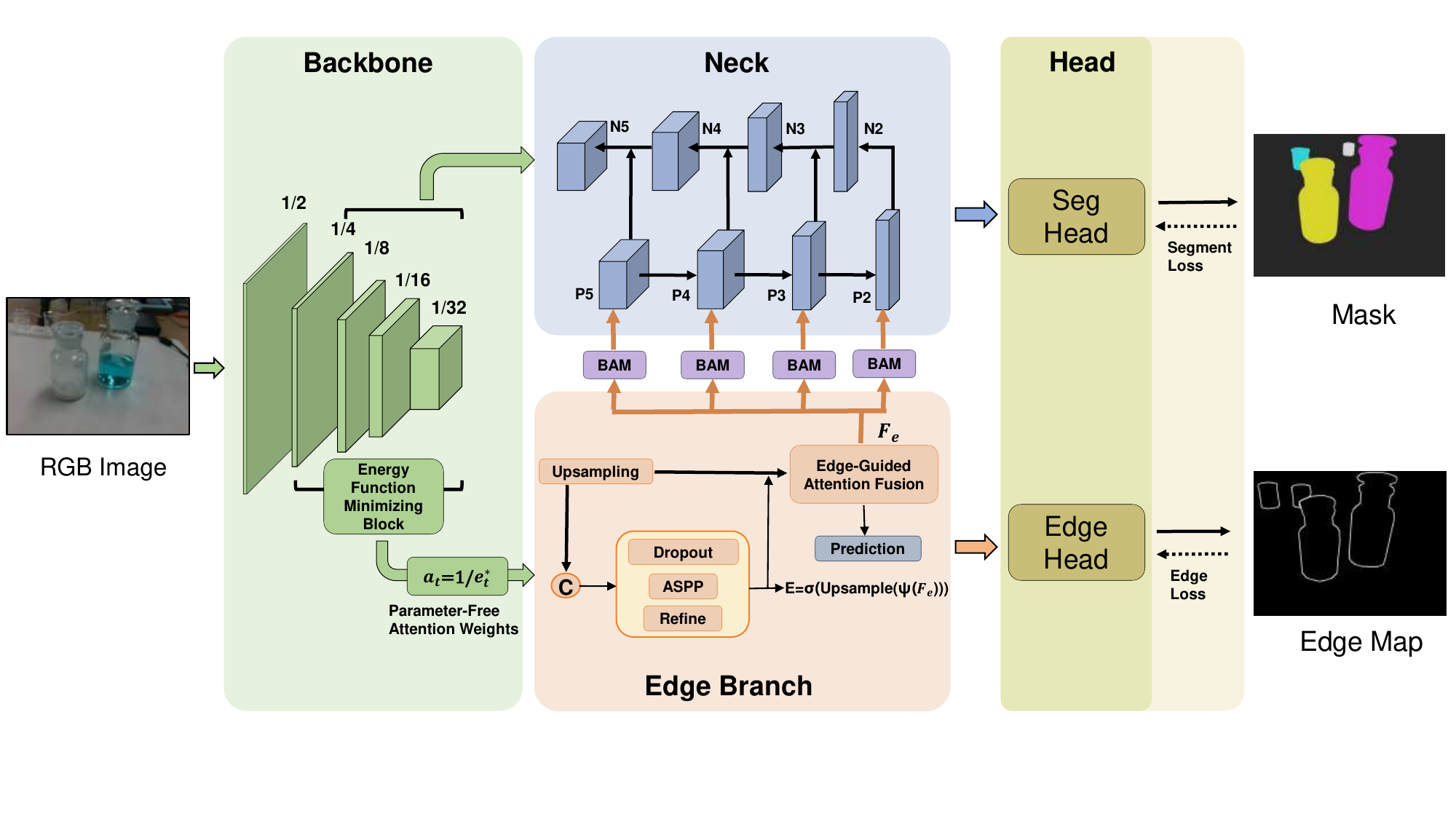}

\vspace{-10mm}

\caption{Overview of the proposed method. It comprises a hierarchical backbone with SimAM attention for multi-scale feature extraction, an auxiliary edge branch for boundary cue enhancement, and a feature fusion neck incorporating Bottleneck Attention Module (BAM). By jointly optimizing instance segmentation and edge detection, the model achieves superior contour precision for transparent objects.}
\label{fig_2}
\end{figure*}

Transparent glass vessels often exhibit unstable interior visual cues, while their boundaries remain relatively reliable. Motivated by this observation, a lightweight multi-scale edge detection module is introduced into the YOLOv5-Seg backbone. This module provides explicit boundary supervision and generates edge-aware features used for subsequent feature fusion in the network neck. The overall framework of the proposed network is illustrated in Fig.~\ref{fig_2}.

\subsubsection{Multi-scale Edge Feature Extraction}

The edge branch takes multi-scale feature maps $\{P_2, P_3, P_4, P_5\}$ from the backbone as inputs. High-resolution features capture fine boundary details, while lower-resolution features provide more stable semantic context. To reduce computational overhead and unify feature representations, feature maps from different scales are first compressed to a common intermediate channel dimension and spatially aligned. They are then upsampled to the same spatial resolution and fused along the channel dimension.

To enlarge the receptive field and capture contextual information, a lightweight Atrous Spatial Pyramid Pooling (ASPP) module is employed. The resulting edge-aware feature representation is denoted as $F_e$, which preserves boundary details while incorporating cross-scale semantic consistency and suppressing background noise. The structure of the edge branch is illustrated in Fig.~\ref{fig_2}.

\subsubsection{Edge Confidence Map Prediction}

Based on the edge-aware feature representation $F_e$, a lightweight prediction head $\psi$ generates an edge confidence map, which is upsampled to the original image resolution:
\begin{align}
    E = \sigma\big(\mathrm{Upsample}(\psi(F_e))\big),
\end{align}
where $E$ denotes the pixel-wise edge confidence and $\sigma(\cdot)$ represents the sigmoid activation function. The edge prediction head is used only for auxiliary supervision during training, while the intermediate edge-aware features are fused into the neck and retained during inference.

\subsubsection{Edge-Guided Attention Fusion in the Neck}

The aligned multi-scale edge features are further fused into the network neck through a Bottleneck Attention Module(BAM). These features do not directly participate in the loss computation; instead, they serve as edge-guided attention inputs that guide the network to emphasize boundary regions during inference.

The three components work in a cooperative manner to enhance transparent object perception. Multi-scale edge feature extraction provides stable boundary representations, while edge confidence prediction introduces explicit supervision to enforce boundary-aware learning. The edge-guided attention fusion further injects these features into the neck to emphasize reliable contours and suppress background noise. Since object boundary prediction and instance segmentation share strong structural consistency, emphasizing boundary regions through this progressive pipeline helps improve segmentation accuracy while suppressing noise caused by refraction effects and background textures.

\subsubsection{SimAM-Based Feature Refinement}

To improve feature discrimination with minimal parameter overhead, 
the SimAM attention module~\cite{simam} is inserted after the C3 
modules. Transparent glassware often exhibits weak texture, refraction, 
and low contrast, causing object responses to be easily confused with 
background regions. SimAM alleviates this problem by assigning 
neuron-wise three-dimensional attention weights according to an 
energy-based measure of feature separability. 

Specifically, SimAM measures the linear separability between a target neuron and other neurons within the same channel and defines the following energy function:
\begin{align}  
\begin{aligned}     
e_t=&\frac{1}{M-1}\sum\nolimits_{i=1}^{M-1}(-1-(w_tx_i+b_t))^2 \\
   &+(1-(w_tt+b_t))^2+\lambda w_t^2,
\end{aligned}
\end{align}
where $t$ denotes the target neuron, $x_i$ represents other neurons in the same channel, and $M=H\times W$ is the number of neurons in the channel. By minimizing this energy, the attention weight is obtained as
\begin{equation}
a_t=1/e_t^*.
\end{equation}

Through this mechanism, SimAM emphasizes informative features and suppresses misleading high-energy responses generated by refraction and specular highlights, thereby improving the representation of transparent glass regions in complex environments.

\begin{figure}[!t]
\centering
\vspace{2mm}      

\begin{subfigure}[t]{0.3\linewidth}
    \centering
    \includegraphics[trim=0 40 0 0,clip,width=\linewidth]{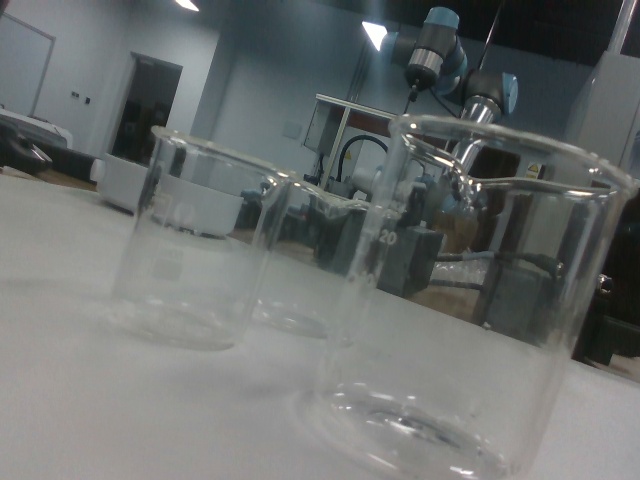}
\end{subfigure}
\begin{subfigure}[t]{0.3\linewidth}
    \centering
    \includegraphics[trim=0 10 0 0,clip,width=\linewidth]{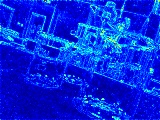}
\end{subfigure}
\begin{subfigure}[t]{0.3\linewidth}
    \centering
    \includegraphics[trim=0 10 0 0,clip,width=\linewidth]{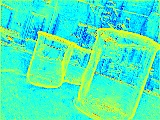}
\end{subfigure}

\vspace{2mm}

\begin{subfigure}[t]{0.3\linewidth}
    \centering
    \includegraphics[trim=0 40 0 0,clip,width=\linewidth]{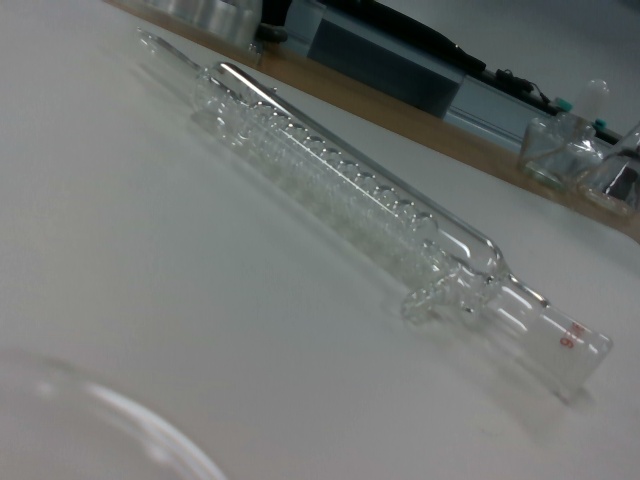}
\end{subfigure}
\begin{subfigure}[t]{0.3\linewidth}
    \centering
    \includegraphics[trim=0 10 0 0,clip,width=\linewidth]{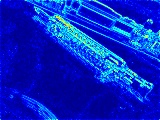}
\end{subfigure}
\begin{subfigure}[t]{0.3\linewidth}
    \centering
    \includegraphics[trim=0 10 0 0,clip,width=\linewidth]{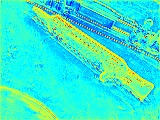}
\end{subfigure}

\caption{From left to right: RGB images, intermediate feature maps of YOLO-Seg, and intermediate feature maps enhanced by the edge module and SimAM attention mechanism.}
\label{fig:edge}
\end{figure}

As illustrated in Fig.~\ref{fig:edge}, the visualization of intermediate feature maps shows that the integration of the edge module and the SimAM attention mechanism leads to clearer object boundaries and significantly suppresses irrelevant surrounding regions.

\subsection{Multi-view 3D Centroid Estimation}

For real-time collision avoidance during robotic manipulation, the proposed system estimates a coarse yet reliable 3D position for each segmented glass vessel using multi-view observations collected during robot motion.

For a detected instance under the $i$-th viewpoint, the 2D geometric centroid is computed directly from the predicted segmentation mask. Let the predicted mask be $\mathcal{M}_i$. The centroid in the image plane is defined as
\begin{align}
\mathbf{c}_i = (u_i, v_i) =
\frac{1}{|\mathcal{M}_i|}
\sum\nolimits_{(u,v)\in \mathcal{M}_i} (u,v).
\end{align}

Compared with bounding box centers, mask centroids more accurately reflect the projected object location and remain stable under partial occlusion, reducing triangulation jitter across views.

Given the camera intrinsic matrix $\textit{K}$, the centroid $\textit{c}_i$, and the camera pose
$\textit{T}^{\text{base}}_{\text{cam},i}=[\textit{R}_i,\textit{C}_i]$, the viewing ray in the camera frame can be computed as
\begin{align}
\textit{d}_i^{\text{cam}} =
\textit{K}^{-1}[u_i\; v_i\; 1]^\top/
\|\textit{K}^{-1}[u_i\; v_i\; 1]^\top\|.
\end{align}
The ray direction in the robot base frame is
\begin{align}
\textit{d}_i = \textit{R}_i \textit{d}_i^{\,\text{cam}}.
\end{align}
Here $\textit{C}_i$ denotes the camera center in the base frame and $\textit{R}_i$ represents the rotation matrix.

Since the robot end-effector equipped with an RGB camera performs small controlled motions, multiple observations $(\textit{C}_i,\textit{d}_i)$ are collected from different viewpoints. Ideally, the 3D centroid $\textit{p}$ lies at the intersection of these rays. Considering inevitable perception noise, a least-squares multi-ray triangulation method is used to estimate the 3D centroid position of the transparent object. The distance from point $\textit{p}$ to the $i$-th ray can be written as
\begin{align}
\left\|(\textit{I}-\textit{d}_i\textit{d}_i^{\top})(\textit{p}-\textit{C}_i)\right\|.
\end{align}
The estimated 3D position is obtained by solving
\begin{align}
\hat{\textit{p}}=
\arg\min_{\textit{p}}
\sum_i
\left\|
(\textit{I}-\textit{d}_i\textit{d}_i^{\top})
(\textit{p}-\textit{C}_i)
\right\|^2.
\end{align}

Although simpler than dense 3D reconstruction, this centroid-based estimation is sufficient to construct conservative bounding volumes for fragile glass vessels and can be efficiently integrated into the robot motion planning pipeline.

\subsection{Perception-Guided Obstacle Modeling}

Based on the estimated 3D centroid of each glass vessel, a conservative bounding volume is constructed to support safe collision avoidance during robotic manipulation.

Given the estimated centroid $\bm{p}^* \in \mathbb{R}^3$, each glass vessel is approximated by an axis-aligned cubic bounding volume centered at $\textit{p}^*$. The cube size is conservatively selected to cover the spatial extent of the object while accounting for segmentation uncertainty and geometric variations, and can be adjusted according to object categories to balance safety and workspace utilization.

During robot motion planning, these bounding volumes are treated as dynamic obstacles in the workspace, and collision checking is performed between robot links and the volumes to generate safe and collision-free trajectories. Meanwhile, the robot continuously maintains multi-view observation of glass vessels and updates their centroids and bounding volumes online, ensuring stable perception and recognition of fragile transparent objects throughout the collision avoidance process. The overall workflow is shown in Algorithm~\ref{alg:reactive-planner}.

\begin{algorithm}[t]
\caption{Real-Time Perception-Guided Obstacle Avoidance}
\label{alg:reactive-planner}
Initialize shared obstacle map $\mathcal{O} \leftarrow \emptyset$ \\
\textbf{Thread 1: Perception and 3D Estimation}\;
\While{camera is running}{
    $\mathcal{D} \leftarrow \text{Seg Model}(\text{camera frame})$\\
    \For{each $d \in \mathcal{D}$}{
        Compute centroid of $d$ \\
        Back-project centroid to a 3D ray using camera matrix $\textit{K}$ and pose $\textit{T}_{cam}$\\
    }
    Update tracks using mask IoU and append observation rays\\
    \If{number of rays $\geq N_{min}$}{
    
        Estimate 3D point $\textit{P}$ by least-squares ray intersection \\
        $\mathcal{O} \leftarrow \mathcal{O} \cup \{\textit{P}\}$\\
    }
}
\textbf{Thread 2: Motion Planning and Execution}\;
\For{each waypoint $\mathbf{w}_i$}{
    Plan trajectory $\pi$ from $\mathbf{q}_{curr}$ to $\mathbf{w}_i$ with obstacle set $\mathcal{O}$\\
    \If{obstacle map updated}{Replan trajectory\\}
    Execute trajectory $\pi$ \\
}
\end{algorithm}

This conservative geometric modeling avoids precise shape reconstruction while ensuring fragile glassware remains protected, reducing computational overhead compared with dense mesh or point cloud representations and making the method suitable for real-time robotic applications.

\section{Experiments}

In this section, we present quantitative and qualitative results to validate our method for transparent glassware segmentation and robotic collision avoidance on real-world LabGlass-IS dataset.

\subsection{Dataset}

To support transparent glassware perception in robotic manipulation scenarios, we construct a dedicated dataset of laboratory glassware with pixel-level instance annotations.

\begin{figure*}[!t] 
\centering 
\includegraphics[trim=5bp 5bp 5bp 5bp, clip,width=5.5in]{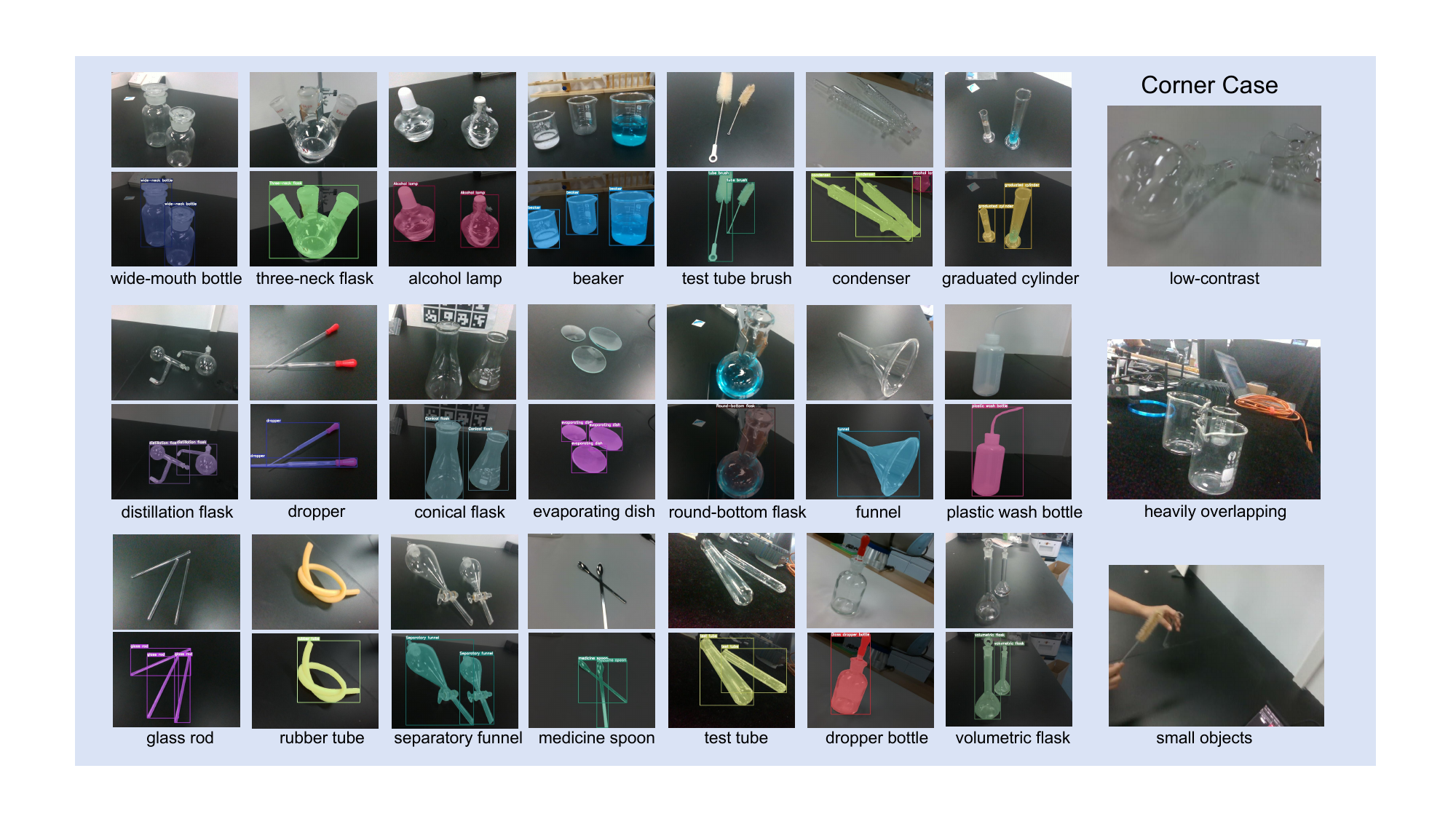} 
\vspace{-4mm}
\caption{Samples of the 21 laboratory apparatus types from our \textit{LabGlass-IS} dataset. The first row shows original images, the second row shows images with detection boxes and instance masks, and the rightmost column presents challenging scenarios.} 
\label{fig_3} 
\end{figure*}

The dataset was captured using the RGB stream of an Intel RealSense D435i camera and contains 3,485 images. All images have a resolution of $640\times480$, with 6,099 annotated instances across 21 categories of common laboratory glassware, covering a wide range of transparent objects in real laboratory environments. The data were collected by recording multi-view, multi-distance videos of glassware on laboratory benches with varying scene complexity, followed by frame extraction and redundancy filtering to remove highly similar frames. These objects exhibit typical transparent characteristics such as refraction, reflection, and weak internal textures. Diverse backgrounds and object arrangements reflect realistic robotic manipulation scenarios. All instances are manually annotated with pixel-level masks, as shown in Fig.~\ref{fig_3}.

\subsection{Implementation Details and Training Settings}

All experiments are implemented using the PyTorch framework and trained on an NVIDIA RTX 4090 GPU. Input images are resized to $640\times640$ during training and inference.

YOLOv5-Seg is adopted as the baseline framework. Without changing its overall architecture, we integrate an edge detection module, an edge-guided attention fusion mechanism, and a parameter-free attention module to enhance boundary perception and feature representation with minimal overhead.

During training, the Adam optimizer is employed with an initial learning rate of $1\times10^{-3}$ and a batch size of 16. To improve model generalization, data augmentation techniques including random horizontal flipping and brightness and contrast perturbations are applied.

The edge confidence map is used only during training as an auxiliary supervision signal to guide the network toward boundary-aware feature learning. No additional computation related to edge supervision is introduced during inference.

\subsection{Experimental Results and Analysis}

\subsubsection{Quantitative Evaluation}

To evaluate the performance of the proposed transparent glassware instance segmentation method, we compare representative segmentation approaches, including one-stage detection models and interactive segmentation methods.

Notably, the FastSAM~\cite{fast} model is also included for comparison. Since FastSAM requires prompts (e.g., points, boxes, or semantic inputs) for specific object segmentation, we use object bounding boxes generated by YOLO as prompts for FastSAM to obtain the final instance masks. All methods are trained and evaluated using the same dataset split and training settings to ensure fair comparison.

Instance segmentation performance is evaluated using the following metrics:

\begin{itemize}

\item \textbf{mAP$_{50}$}/\textbf{mAP$_{75}$}: Mean Average Precision computed at IoU thresholds of $0.50$/$0.75$.

\item \textbf{mAP$_{50:95}$}: Mean Average Precision averaged across IoU thresholds from $0.50$ to $0.95$ with a step of $0.05$, providing a stricter evaluation of segmentation quality.

\item \textbf{Boundary F-score (BF)}: A metric specifically measuring how well the predicted mask contours align with the ground truth boundaries within a small distance threshold $\theta$. We set $\theta = 2$ pixels in our experiment.

\item \textbf{BF of Slender Objects}: The Boundary F-score calculated specifically on a subset of the dataset containing thin and elongated glassware (glass rods and droppers), which are particularly challenging due to severe refraction and weak visual cues.

\end{itemize}

Table~\ref{tab1} presents the quantitative comparison results on the test set. The results show noticeable performance differences among general-purpose segmentation models in transparent glassware scenarios due to reflection, refraction, and transparency effects.
\begin{figure*}[!t]
\centering
\includegraphics[width=150mm,height=85mm]{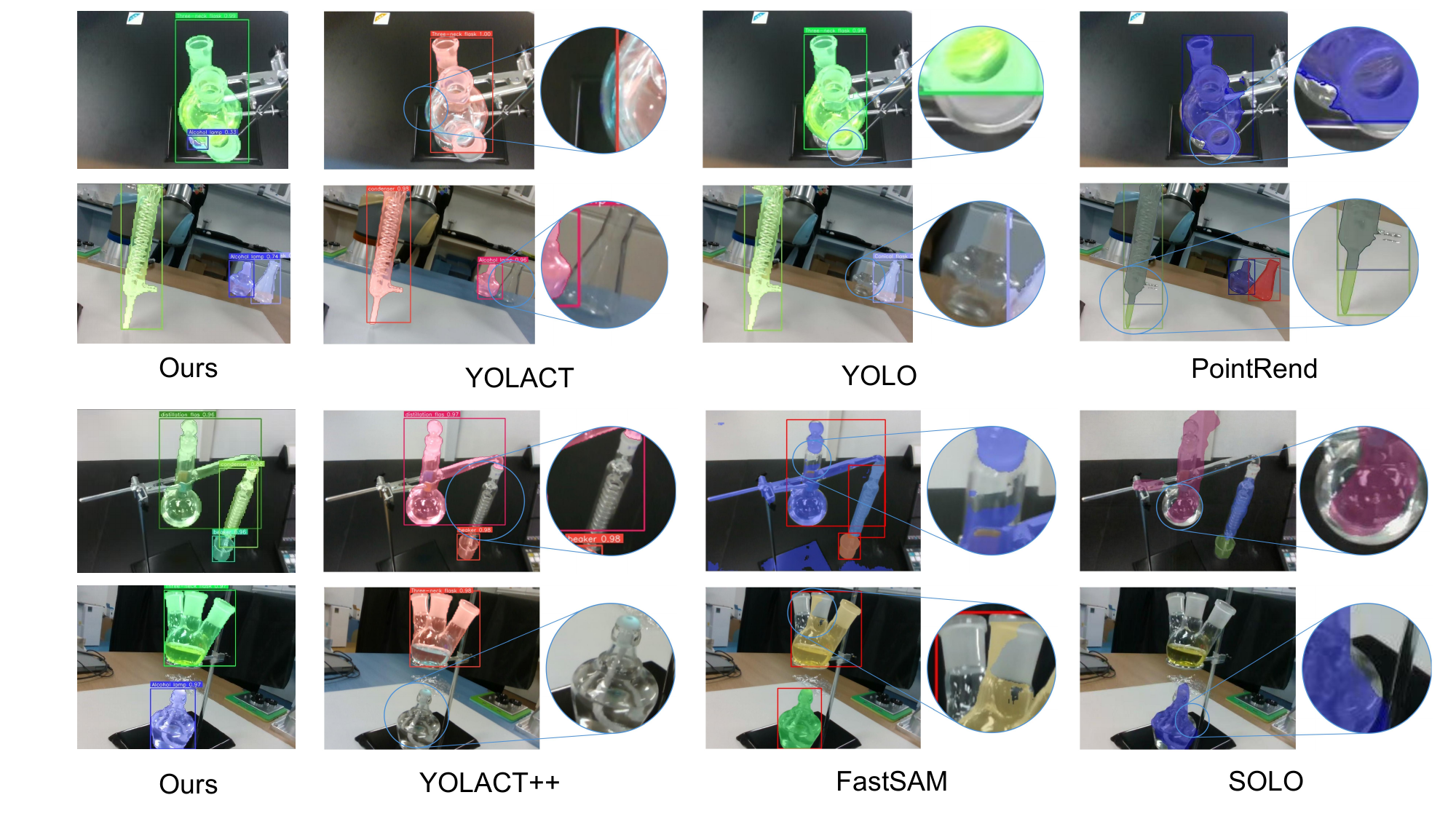}

\vspace{-3mm}

\caption{Qualitative comparison on transparent glassware instance segmentation. Zoomed-in regions mainly highlight the segmentation errors of other methods, including missed small objects, boundary misalignment, incomplete masks and over-segmentation of complex glassware, while the proposed method produces more complete and accurate masks with clearer boundaries and better preservation of slender structures.}
\label{fig_4}
\end{figure*}

\renewcommand{\arraystretch}{1.2}
\setlength{\tabcolsep}{4pt}
\begin{table*}[t]
\centering
\caption{Quantitative Comparison on the Test Set}
\label{tab1}
\begin{tabular}{l c c c c c c c c}
\toprule
Method &Boundary F-score (BF) $\uparrow$ &BF of Slender Objects $\uparrow$ &mAP$_{50}\uparrow$ & mAP$_{75}\uparrow$ & mAP$_{50:95}\uparrow$ &FPS & Time (ms) &Params (M) \\
\midrule
SOLO &50.06 &56.23 &71.80 &42.10 &42.90  &9.82  &101.80  &276.21  \\
YOLO+FastSAM &78.87 &67.69 &68.20  &55.60 &52.10  &6.25  &160.10  &144.06  \\
YOLACT    &89.93 &88.73  &95.54  &80.72 &68.60  &43.50  &22.99  &190.35  \\
YOLACT++  &89.74 &89.49  &88.61  &84.53 &77.38  &39.08  &25.59  &196.44  \\
PointRend &88.91 &88.91  &98.03  &92.32 &\textbf{84.35}  &50.00  &20.02  &603.69  \\

\hline
Ours &\textbf{97.80} &\textbf{98.21}  &\textbf{98.30}  &\textbf{92.90} &82.20  &\textbf{140.85}  &\textbf{7.10}  &\textbf{17.18}  \\
\bottomrule
\end{tabular}
\end{table*}

SOLO shows relatively limited performance, while prototype-based methods such as YOLACT and YOLACT++ achieve improved results. PointRend\cite{pointrend} demonstrates strong boundary refinement capability and achieves the highest $mAP_{50:95}$ of 84.35, though its heavy architecture and lower Boundary F-score (88.91) limit its suitability for real-time robotic applications. The combination of YOLO and FastSAM achieves an $mAP_{50:95}$ of 52.1, indicating adaptation challenges of large pretrained models in transparent glassware scenarios.

The proposed method achieves a competitive $mAP_{50:95}$ of 82.2 while delivering a superior BF of 97.80 at a high inference speed (7.1\,ms). Notably, for slender objects (glass rods and droppers), our approach maintains a high BF of 98.21, whereas other models show a marked decline due to refractive interference. Although PointRend achieves a higher $mAP_{50:95}$, the proposed method offers a balance between accuracy and efficiency, which is critical for deployment on resource-constrained robotic platforms.

\subsubsection{Qualitative Evaluation}

Qualitative comparison results are shown in Fig.~\ref{fig_4}. The proposed method demonstrates superior performance in scenarios involving small targets and irregular object shapes.

As illustrated, the original YOLO and YOLACT models frequently miss small targets in the scene and suffer from segmentation errors where bounding boxes and masks fail to align precisely at object boundaries. While YOLACT++ offers marginal improvements, it remains prone to neglecting slender structures or smaller objects. Furthermore, PointRend occasionally splits complex objects such as three-neck flasks into multiple objects, while FastSAM tends to generate incomplete masks. The SOLO model generally exhibits low boundary segmentation accuracy. It is worth noting that even when provided with accurate bounding box prompts, FastSAM sometimes fails to segment the correct object, likely because transparent chemical glassware is underrepresented in its pretraining data.

Overall, the proposed method provides more accurate and stable instance segmentation results in complex transparent object scenarios.

\subsection{Ablation Studies}

To analyze the contributions of the proposed modules, ablation experiments are conducted on the laboratory glassware dataset. All models are trained and evaluated under identical settings, using the original YOLOv5-Seg as the base model while incrementally adding different network components. The results are summarized in Table~\ref{tab2}.

\begin{table}[!t]
\centering
\caption{Ablation Experiments on Our Modules}
\vspace{-0.2cm}
\label{tab2}
\begin{tabularx}{\columnwidth}{l c c c c c c}
\toprule
Method    & Boundary F-score (BF) $\uparrow$ & mAP$_{50}\uparrow$ & mAP$_{50:95}\uparrow$ & mIoU$\uparrow$  \\ \midrule
Baseline  &96.67  &98.1   & 81.3   &88.95                 \\
+SimAM &97.58  &98.1    &81.2    &88.98              \\
+P2    &97.60 &98.0    &81.6    &89.06                  \\
+P2+edge &97.71 &98.0    &81.8    &89.20                 \\
\hline
Ours &97.80 &98.3    &82.2    &89.33                \\ 
\bottomrule
\end{tabularx}
\end{table}

Adding SimAM alone to the baseline substantially improves the Boundary F-score from 96.67 to 97.58, while yielding only a marginal change in $mAP_{50:95}$ (81.3 $\to$ 81.2) and slightly improving mIoU (88.95 $\to$ 88.98). This indicates that SimAM mainly enhances boundary feature discrimination, although its contribution to overall mask accuracy is limited when applied alone.

Introducing the high-resolution P2 feature layer further improves the Boundary F-score to 97.60, while increasing $mAP_{50:95}$ from 81.2 to 81.6 and mIoU from 88.98 to 89.06. This indicates that high-resolution features help preserve fine spatial and boundary details, which is particularly beneficial for slender glassware such as glass rods and droppers.

Adding the lightweight edge prediction branch further improves the Boundary F-score, $mAP_{50:95}$, and mIoU to 97.71, 81.8, and 89.20, respectively, demonstrating that explicit boundary supervision enhances boundary localization and mask completeness.

Overall, the ablation results show that the proposed modules contribute complementary improvements. In particular, high-resolution feature fusion and edge-aware design play a key role in improving segmentation quality.

\subsection{Collision Avoidance}

To evaluate the effectiveness of the proposed perception-driven collision avoidance strategy, the estimated 3D centroids and conservative bounding volumes are deployed on a robotic manipulator in a real lab environment. 

We also measure the latency from object detection to 3D centroid estimation, defined as the time between the first image detection and the availability of its 3D centroid. The average per-frame latency is 73.70 ms, and the average time to obtain the first valid 3D centroid is 290.50 ms, demonstrating the system's suitability for real-time collision avoidance.

\textbf{3D centroid estimation accuracy.}
We evaluate 5 scenes with 1--5 glass vessels. Ground-truth centroids are measured manually ($\pm5$\,mm precision) in the robot base frame. Table~\ref{tab:3d_accuracy} reports the Euclidean errors, with an average of 38.0\,mm and a maximum of 60.5\,mm. Errors are mainly along the depth ($z$) axis due to limited triangulation resolution. Critically, the conservative bounding volumes are sized to fully enclose the objects with this margin, so such errors do not compromise safety.

\begin{table}[!t]
\centering
\caption{3D Centroid Estimation Accuracy Across Different Scenes}
\vspace{-0.2cm}
\label{tab:3d_accuracy}
\begin{tabularx}{\columnwidth}{l c c c c}
\toprule
Scene & \#Objects & Mean Error (mm) & Max Error (mm) & \#Views \\
\midrule
S1    & 1  & 25.3 & 25.3 & 27 \\
S2    & 2  & 40.2 & 45.1 & 27 \\
S3    & 3  & 37.5 & 45.1 & 21 \\
S4    & 4  & 50.3 & 60.5 & 27 \\
S5    & 5  & 30.1 & 40.4 & 27 \\
\hline
\textbf{Avg} & -- & \textbf{38.0} & \textbf{60.5} & -- \\
\bottomrule
\end{tabularx}
\end{table}

\textbf{Collision avoidance success rate.}
We conduct 15 trials across three difficulty levels (Easy/Medium/Hard). A trial succeeds if the robot reaches the goal without contact. As shown in Table~\ref{tab:avoidance_rate}, the system achieves a 93.3\% success rate. The average MoveIt re-planning time is below 20\,ms, confirming that the perception-to-planning pipeline introduces negligible overhead at each update cycle.

\begin{table}[!t]
\centering
\caption{Obstacle Avoidance Success Rate Under 3 Difficulty Levels}
\vspace{-0.2cm}
\label{tab:avoidance_rate}
\begin{tabularx}{\columnwidth}{l c c c c}
\toprule
Difficulty & \#Trials & \#Objects & Success Rate & Avg Plan Time (s) \\
\midrule
Easy   & 5 & 1 & 5/5 & 0.017 \\
Medium & 5 & 2 & 5/5 & 0.018 \\
Hard   & 5 & 3 & 4/5 & 0.018 \\
\hline
\textbf{Total} & \textbf{15} & -- & \textbf{14/15} & \textbf{0.017} \\
\bottomrule
\end{tabularx}
\end{table}

\begin{figure}[!t]
\vspace{2mm}
\centering
\includegraphics[width=170mm,height=100mm]{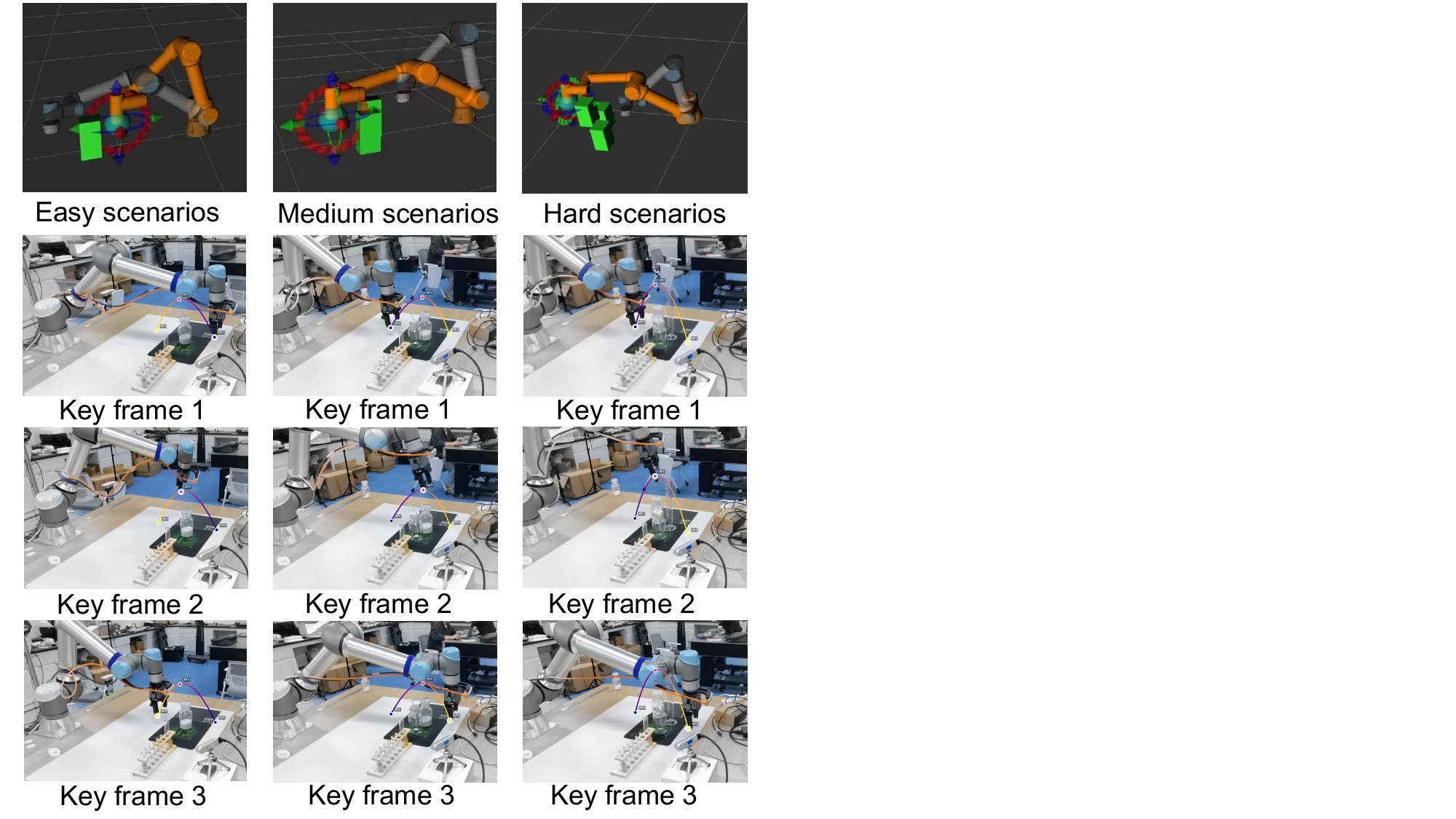}

\vspace{-1mm}

\caption{Real-robot obstacle avoidance experiment. The red curve represents the planned obstacle avoidance trajectory. Columns show Easy/Medium/Hard scenarios; rows show key frames from start, avoidance, to goal.}
\label{fig:glass_result}
\end{figure}

\textbf{Qualitative results.}
Fig.~\ref{fig:glass_result} shows representative keyframes of the avoidance process, where the robot detects obstacles, plans a collision-free trajectory, and reaches the goal safely.

\section{Conclusions}

Experimental results show that the proposed method achieves an effective balance between segmentation quality, real-time efficiency, and robotic usability for transparent laboratory glassware. On LabGlass-IS, it attains 82.2 $\text{mAP}_{50:95}$ at 7.1\,ms per frame, while the real-robot system reaches a 93.3\% collision avoidance success rate. These results indicate that explicit edge-aware design, combined with multi-view centroid estimation and conservative obstacle modeling, provides a practical perception-to-action solution for robot collision avoidance in cluttered transparent-object scenes.

The current system is primarily designed for reliable collision avoidance and does not yet support more precise geometric understanding or fine manipulation. Future work will focus on improving geometric perception accuracy and enabling more precise interaction capabilities, thereby extending the framework to more complex manipulation tasks.

\bibliographystyle{IEEEtran}
\bibliography{reference}

\end{document}